\documentclass[11pt]{article}

\usepackage[margin=1.1in]{geometry}
\usepackage[T1]{fontenc}
\usepackage[utf8]{inputenc}
\usepackage{lmodern}
\usepackage{microtype}
\usepackage{graphicx}
\usepackage{booktabs}
\usepackage{amsmath}
\usepackage[numbers,sort&compress]{natbib}
\usepackage{xcolor}
\usepackage[colorlinks=true,linkcolor=blue!60!black,citecolor=blue!60!black,urlcolor=blue!60!black]{hyperref}
\usepackage{caption}
\usepackage{multicol}
\newcommand{\eg}{e.g.,\ }

\title{\textbf{Structured Output Collapses\\Answer Diversity Across 44 Language Models}}
\author{Tapan Parikh\\Cornell Tech\\\texttt{tsp53@cornell.edu}}
\date{July 2026}

\begin{document}
\maketitle

\begin{abstract}
\noindent
When a language model must choose one answer from a large space of equally valid options, a
format clause --- \emph{``Reply with JSON only''} --- changes which answer it chooses. We re-run
the One-Word Census \citep{parikh2026census}, 31 category prompts with wide answer spaces
asked of 44 models, with the reply requested in JSON: no schema enforcement, no constrained
decoding, only the request. The field's convergence deepens sharply: on the unconstrained
prompt (``Pick a word.''), the modal answer's share rises from 41\% to 64\% and distinct
answers fall from 52 to 36; mean answer-choice surprisal drops 1.80 to 1.58 bits. The tax is
\emph{progressive} and respects the instrument's resolution: six of 44 models move
individually (BH-FDR $q=.10$), all toward the mode, led by the panel's most distinctive
frontier model (the strongest explorer loses 1.31 bits --- half its measured
distinctiveness). The format is a sharpener, not a re-indexer: the plain-chat modal answer
survives in 28 of 31 categories, the three flips landing where the plain mode was weakest.
\emph{Defaults are register-indexed}: a within-run re-sample ($n{=}20$) finds JSON shifts 53\%
of a model's stable chat defaults, mostly back to the crowd, and \emph{installs} defaults absent
from chat (Claude Fable 5 answers \emph{cerulean} for colour 0\% of the time in chat, 100\% in
JSON). Full-battery controls reveal a register gradient: compression is significant and specific
to the answer-delivery formats models are trained to speak (JSON $-0.22$ bits, $p=.0002$;
XML $-0.19$, $p=.002$; permutation over models and categories), absent for YAML and CSV, and
\emph{reversed} for an arbitrary bracket wrapper ($+0.13$, $p=.009$) --- weighing the mechanism
toward tool-use post-training, though every serialization concentrates the unconstrained pool, so
a corpus-register component remains. Enforcing the schema at the decoder
(\texttt{response\_format}) compresses no further than the request ($-0.03$ bits): the collapse
lives in the model's response to the register, not the decoder. Structured output is
how software consumes language models; that surface is served by a measurably more homogeneous
model than the chat surface on which models are evaluated, compared, and chosen.
\end{abstract}

\section{Introduction}
\label{sec:intro}

Software does not consume language models in prose. It asks for JSON. Every agent that calls a
tool, every pipeline that extracts a field, classifies a record, fills a schema, or routes a
request receives the model's answer inside a data structure --- and that traffic increasingly
dwarfs the chat window in which models are benchmarked, ranked, and chosen. This paper asks a
question about that surface: when the same model answers the same question in a serialization
format instead of prose, does it give the same answer?

The companion One-Word Census \citep{parikh2026census} built a deliberately narrow instrument.
When a model must choose one answer from a large space of equally valid options ---
\emph{``Name a tree. Reply with one word only.''} --- which does it choose, and how often is it
the same answer the rest of the field chooses? The metric is \emph{answer-choice surprisal}: a
leave-one-out measure, in bits, of how unlikely a model's answers are under the pooled answers
of every other model, computed by exact match on normalized one-word replies --- no embeddings,
no LLM judge, no human annotation. Across 44 models it found a
monoculture (\emph{oak} takes 94\% of tree answers; \emph{serendipity} 41\% of answers to an
unconstrained ``Pick a word.'') structured by a wide per-model conformity range and, for a
handful of models, stable off-modal defaults that constitute a measurable character.

We re-run that instrument unchanged in every respect but one: the reply is requested in a
serialization format. \emph{``Name a tree. Reply with JSON only, in the form
\{"word": "<your answer>"\}.''} No schema enforcement, no constrained decoding, no token masking
--- only the request. The 31 prompts, the 44-model panel, and the conditions (no system prompt,
requested temperature 1.0, four samples) are identical to the census; the single manipulated
variable is the register in which the answer is delivered. Because surprisal is computed
\emph{within} each format column --- JSON answers scored against the JSON field --- a column's
convergence is an internal property of that column, not an artifact of comparing one register
against another.

Our contributions:

\begin{enumerate}
\item \textbf{A progressive format tax} (\S\ref{sec:progressive}). Field-mean surprisal falls
from 1.80 to 1.58 bits ($-0.22$ per model, $p=.0002$), but the drop is not uniform: six of
44 models move individually (BH-FDR $q=.10$), all toward the mode, led by the panel's most
distinctive models (the strongest explorer loses 1.31 bits),
while the conformist floor is immobile and part of the positive-$\Delta$ tail is
register-invariant stranding rather than divergence --- a panel-free check separates the two.
The remaining deltas form a noise plateau we decline to rank.
\item \textbf{A sharpener, not a re-indexer} (\S\ref{sec:sharpener}). The plain-chat modal
answer survives in 28 of 31 categories; the three flips occur exactly where the plain mode was
weakest.
\item \textbf{Register-indexed defaults} (\S\ref{sec:defaults}). A within-run re-sample
($n{=}20$) shows the register significantly shifts 53\% of a model's stable chat defaults (mostly
back to the crowd) and \emph{installs} register-only defaults absent from chat --- read from
discrete answer behavior, not noisy score deltas.
\item \textbf{A register gradient} (\S\ref{sec:gradient}). Compression is significant and
specific to the answer-delivery formats models are trained to speak (JSON, XML), absent for
YAML and CSV, and \emph{reversed} for an arbitrary non-data wrapper --- weighing the mechanism
toward tool-use post-training while leaving a corpus-register component visible.
\item \textbf{A cheap, reusable method}. The manipulation is one extra column on a frozen
instrument, and format compliance falls out of every run as a standing capability metric.
\end{enumerate}

\section{Related work}
\label{sec:related}

\paragraph{Format restrictions and accuracy.}
The closest prior work asks whether format \emph{restrictions} degrade task performance.
\citet{tam2024speakfreely} report that requiring structured output lowers reasoning accuracy;
\citet{kurt2024saywhatyoumean} contest the result, attributing it to prompt design and parsing
rather than the format itself. Our question is orthogonal to this debate: our prompts have no
wrong answers, so there is no accuracy to degrade. We measure which valid answer is chosen, not
whether a correct one is reached.

\paragraph{Decoder-level enforcement.}
A 2026 line names the cost of \emph{enforcing} structure at the decoder --- the ``constraint
tax'' \citep{constrainttax2026validity, constrainttax2026tools} and ``format tax''
\citep{formattax2026}: grammar-constrained decoding raises validity while lowering correctness
and suppressing behaviors such as tool calls. But they constrain the sampler with
token masking; we mask nothing and only ask. The effect we measure therefore lives in the
model's response to a register, not in a decoding constraint.

\paragraph{Diversity within grammars.}
\citet{automata2025diverse} engineer diversity back into constrained generation with
automata-based steering. That work presupposes exactly the collapse we characterize --- it is
an engineering fix for a phenomenon the present paper measures at the level of a mere request,
with no grammar in the loop.

\paragraph{Surface form and register.}
\citet{sclar2024formatspread} show model accuracy is sensitive to spurious formatting choices;
the census showed answer \emph{choice} is sensitive to wording (\emph{gemstone} vs.\
\emph{precious stone}). This paper adds a third axis: not how the question is worded but the
register in which the answer is requested. Underlying all of this is the mode-collapse
literature \citep{jiang2025hivemind, zhang2025noveltybench, kirk2024rlhf}, which documents the
convergence we exploit and notes that LLM judges systematically prefer modal outputs --- one
reason our instrument stays judge-free.

\section{Method}
\label{sec:method}

\subsection{Instrument}
The instrument is the \emph{one-word census} unchanged \citep{parikh2026census}: 31 single-turn prompts (30
category prompts of the form \emph{``Name a[n] $X$''} plus one unconstrained \emph{``Pick a
word''}), no system prompt, requested temperature 1.0, four samples per cell. Each answer is
scored by its add-one-smoothed leave-one-out answer-choice surprisal against the pooled answers
of the other 43 models, and a model's headline score is the mean over its valid answers, in
bits. Replies are reduced to a single normalized token by exact-match rules and a mechanical
junk guard (\S\ref{sec:hygiene}).

Surprisal is computed \emph{within} each format column. JSON answers are scored against the JSON field, plain answers
against the plain field. A column's convergence is therefore an internal property of that
column --- it is never a comparison of JSON text against a chat reference, and so cannot be an
artifact of the register shift itself. The per-model \emph{$\Delta$-surprisal} (JSON minus
plain) is the headline quantity: how many bits of a model's distinctiveness the register
removes.

\subsection{Format columns}
The plain-chat column is the census transcripts. Each of five format columns re-runs the full
31-category battery for all 44 models at four samples per cell (5{,}456 calls per column):
JSON, XML, YAML, CSV, and a non-data ``brackets'' wrapper, the last included to separate
serialization from mere structure. The clauses are appended verbatim to the census prompt:

\begin{itemize}
\item \textbf{json}: \texttt{Reply with JSON only, in the form \{"word": "<your answer>"\}.}
\item \textbf{xml}: \texttt{Reply with XML only, in the form <word>your answer</word>.}
\item \textbf{yaml}: \texttt{Reply with YAML only, in the form `word: <your answer>`.}
\item \textbf{csv}: \texttt{Reply with CSV only: a header row `word`, then one data row with your answer.}
\item \textbf{brackets}: \texttt{Reply with your answer inside square brackets only, like [answer].}
\end{itemize}

No \texttt{response\_format} parameter, no tool schema, and no constrained decoding is used
anywhere: every column is an ordinary chat completion whose only difference from the census is
the appended clause. This is a deliberate scope choice: these columns measure the effect of the
\emph{request}. An engine-enforced \texttt{response\_format} column is analyzed separately in
\S\ref{sec:enforced}.

\subsection{Parsing and hygiene}
\label{sec:hygiene}
Each reply is reduced to a single answer token in two steps. First the format wrapper is
stripped by a per-format regular expression --- the contents of the JSON \texttt{"word"} field,
the text between \texttt{<word>} tags, and so on --- and the extracted string is then passed
through the census normalization and junk guard unchanged. A reply that does not match its
format's wrapper falls back to the census rule applied to the raw text, so that a malformed
reply cannot smuggle in a spurious ``novel'' answer: the last word of a broken-JSON essay is
scored exactly as the census would score it. One guard is added beyond the census rule, because
a fill-in wrapper invites an artifact the census prompt cannot: a reply whose token is the
category noun or the template placeholder --- a literal \texttt{[city]} for ``name a city'', or
\texttt{answer}/\texttt{word} lifted from the slot --- is
dropped as a failed cell. Parse-plus-junk survival is at least 99\% per
model per column, and we audited per-model survival for every model whose surprisal
\emph{rose}, since non-compliance is the one artifact that manufactures apparent divergence
(\S\ref{sec:gradient}).

Of the 27{,}280 format cells, 91 (0.3\%) are unrecoverable nulls, concentrated in YAML (56) and
negligible in JSON (1); all distributional claims are computed over the surviving cells, with
the per-column parse survival audited above.

\section{Results}
\label{sec:results}

\subsection{The format tax is progressive}
\label{sec:progressive}
\begin{figure}[tp]
\centering
\includegraphics[width=0.62\linewidth]{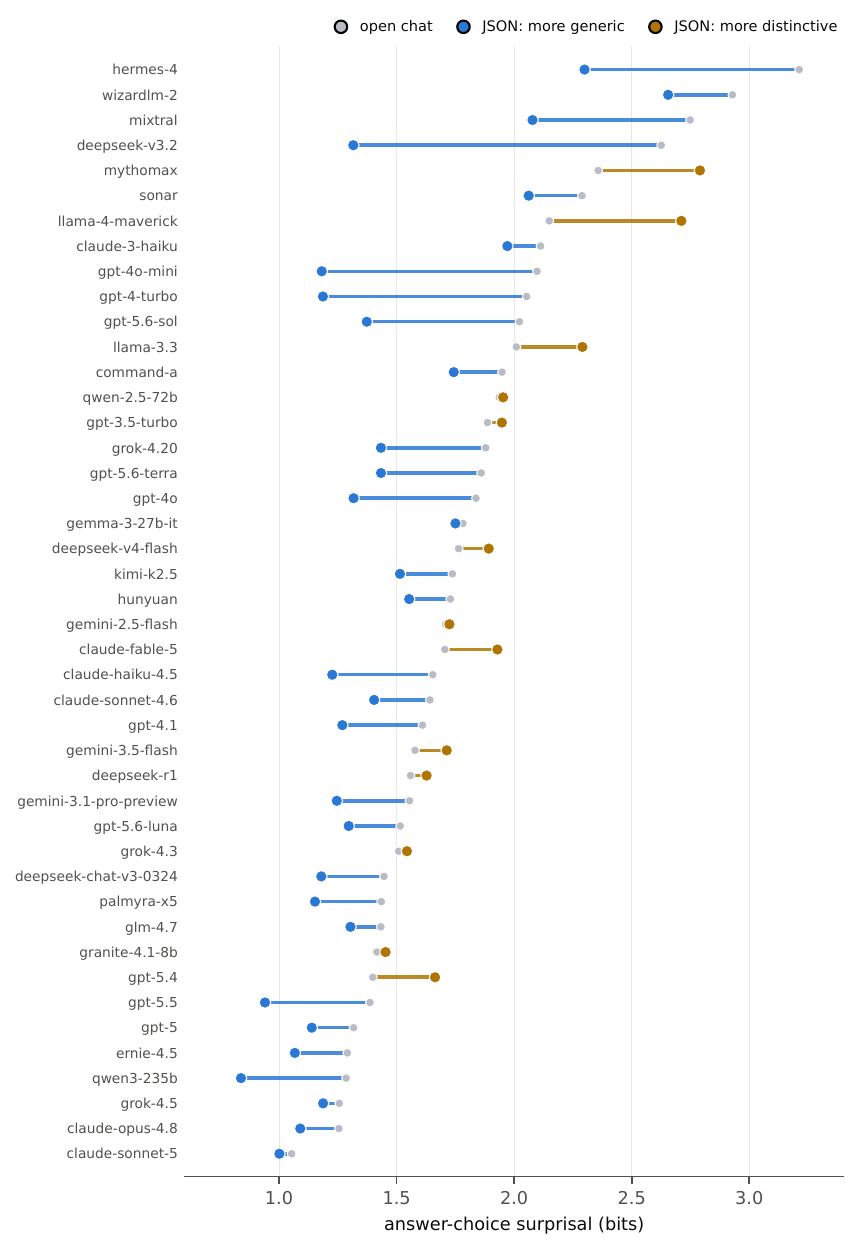}
\caption{Answer-choice surprisal for all 44 models, open chat (gray dot)
versus JSON (colored dot), sorted by the open-chat score. Blue marks a model that becomes
\emph{more generic} under JSON (31 of 44); amber marks one whose relative score \emph{rises}
(13 --- mostly register-invariant models the collapsing field strands, \S\ref{sec:gradient}).
The distinctive models at the top
slide farthest toward the conformist floor, which is itself immobile, and the field mean falls
from 1.80 to 1.58 bits. The raw range is nearly unchanged ($2.2 \to 2.0$) because the
amber models hold up the top: the compression is in the mean and the tail, not the extremes.}
\label{fig:openvsjson}
\end{figure}

Field-mean answer-choice surprisal falls from 1.80 bits in plain chat to 1.58 in JSON. Averaged
over the 44 models the per-model $\Delta$ is $-0.22$ bits, significant on both exchangeable
units (\S\ref{sec:gradient}), but the effect is concentrated rather than spread. The shape is a divergent tail collapsing onto a
fixed floor (Figure~\ref{fig:openvsjson}).

The models that lose the most are the ones that had the most to lose. \textsc{deepseek-v3.2},
the census's one genuine explorer among frontier-lab models, falls from 2.63 to 1.32 bits
($-1.31$) --- half its measured distinctiveness --- the moment the answer is requested as JSON.
\textsc{hermes-4} loses 0.91, \textsc{gpt-4o-mini} 0.92, \textsc{gpt-4-turbo} 0.87,
\textsc{gpt-5.6-sol} 0.65. The conformist floor barely moves: \textsc{claude-opus-4.8} $-0.16$,
\textsc{grok-4.5} $-0.07$, \textsc{claude-sonnet-5} $-0.05$. There is nowhere below the floor to
go, and the tail falls toward it, though the raw range is nearly unchanged (2.2 to 2.0 bits): a
register-invariant minority holds its ground while the field collapses around it
(\S\ref{sec:gradient}),
so the compression is in the mean and the collapsing tail, not the extremes. On the
unconstrained ``Pick a word'' prompt the same collapse appears with no category to
anchor it: \emph{serendipity} rises from 41\% of the pool to 64\% and the number of distinct
words falls from 52 to 36.

Which individual movements are real we take up in \S\ref{sec:gradient}; the progressive
\emph{shape} --- tail down, floor fixed --- is a property of the aggregate and does not depend
on ranking the middle of the table. Nor is the shape regression to the mean, the classic
alternative for ``the ones that lost most had the most to lose'': split-half test-retest
reliability of the census score is $r{=}0.94$, so the shrinkage a pure re-measurement predicts
for the most distinctive model is $0.05$ bits against the $1.31$ observed, and every
individually significant compressor (\S\ref{sec:gradient}) clears the re-measurement-noise null
by $3.7$ to $10\sigma$.

\subsection{A sharpener, not a re-indexer}
\label{sec:sharpener}
If the register merely added noise, or merely imposed structure, it could move probability
anywhere. Instead it sharpens the distribution the model already had. In 28 of 31 categories the
JSON modal answer is the same word as the plain-chat mode, and less than 4\% of JSON
answer-mass lands on words the plain census never produced. The register moves mass
\emph{onto the existing mode}, not onto new answers.

The three exceptions are diagnostic: each is a category whose plain-chat mode was weak. Insect
flips from \emph{ant} (43\%) to \emph{butterfly} (60\%), board game from \emph{chess} to
\emph{monopoly}, dance from \emph{salsa} to \emph{tango}. The register amplifies its own
conditional prior, and that prior wins only where the chat prototype was too weak to hold.

\subsection{The register moves the mode, not the temperature}
\label{sec:temperature}
A natural worry is that we are measuring a sampling change rather than a change in preference ---
that some providers quietly cool the sampler when a request looks like structured output. The
census's self-distinctness statistic (distinct answers divided by samples, within model and
category) is the effective-temperature proxy that settles it. At the field level it is nearly
flat across all six columns --- plain 0.42, JSON 0.39, XML 0.40, YAML 0.42, CSV 0.43, brackets
0.43 --- while surprisal drops 0.22 bits under JSON. A serving-layer temperature cut would
crater self-distinctness wherever it cut; it does not. The collapse is \emph{positional}: mass
relocates onto the field's mode without the model sampling any less around it.

The within-model narrowing that does exist is progressive, on the same models as the surprisal
effect: \textsc{hermes} $0.71\to0.52$, \textsc{deepseek-v3.2} $0.56\to0.42$, \textsc{sonar}
$0.50\to0.40$, \textsc{haiku-4.5} $0.42\to0.33$, while the conformist floor is flat
(\textsc{sonnet-5} $0.30\to0.27$) and \textsc{claude-fable-5} holds at 0.32 in both registers ---
stable self-distinctness rather than a cooled sampler. The tax is thus progressive on both of the
census's axes, and the dissociation between a large surprisal move and a small self-distinctness
move is itself the signature of a reshaped conditional distribution rather than a rescaled one.

\subsection{Defaults are register-indexed}
\label{sec:defaults}
The sharpest evidence that personality is register-dependent comes not from surprisal scores,
which are noisy for individual models (\S\ref{sec:gradient}), but from discrete answer behavior.
Call a model's \emph{stable default} in a category an off-modal answer it gives repeatedly --- a
standing refusal of the field's first choice, like \textsc{fable}'s \emph{gouda} for cheese. The
four-sample census flags 144 such defaults across the panel. Four samples, however, cannot tell a
genuine default from a lucky streak, nor a register effect from ordinary run-to-run drift, so we
re-sampled every default cell at $n=20$ in \emph{both} registers within a single run. This
settles three questions the four-sample data could not.

\paragraph{The defaults are real, and the register erodes them.} At $n=20$ the flagged defaults
are genuine (median per-sample probability $0.90$), not sampling artifacts. Requesting JSON then
\emph{significantly} shifts the answer distribution for 76 of 144 (53\%) of them (Fisher exact
per cell, Benjamini--Hochberg $q=.10$), against a false-positive floor of ${\sim}10\%$, and the
shift is directional: 29\% revert outright to the field's modal answer. The register pulls a
model's idiosyncratic default back toward the crowd.

\paragraph{The register also installs defaults.} The move is not only subtractive. Of the JSON
answers a model gives four-of-four but \emph{never} produces in chat, 81\% survive the $n=20$
re-sample as genuine register-only defaults: \textsc{fable} says \emph{cerulean} for colour 0\%
of the time in chat and 100\% in JSON ($p\approx10^{-11}$), and \emph{carpenter} for occupation
$0\%\to90\%$ ($p\approx10^{-9}$); \textsc{opus-4.8} acquires \emph{phoenix}, \textsc{sonnet-4.6}
\emph{gold}. Some of the field-level mode flips of \S\ref{sec:sharpener} are these acquisitions
in aggregate --- one model's JSON-only \emph{butterfly} ($15\%\to100\%$) or \emph{monopoly}
($5\%\to100\%$).

\paragraph{Same default, opposite response.} Because the register both erases and installs,
models that look identical in chat diverge under it. \textsc{gpt-5.6-terra} and \textsc{fable}
both answer \emph{mango} for fruit, twenty runs of twenty in chat; under JSON \textsc{terra} flips
to \emph{apple} (20/20, $p\approx10^{-11}$) while \textsc{fable} holds \emph{mango} (19/20). The
right object is therefore not one personality the register reveals or hides but a
\emph{per-register defaults profile}: a model's character is indexed to the channel it is asked
through, the same way the census found answers indexed to the wording of the question.

\subsection{Serialization vs.\ structure: the register gradient}
\label{sec:gradient}

\paragraph{The gradient.}
Extending the same battery to four further formats separates \emph{serialization} from mere
\emph{structure}, and the result is a gradient rather than a uniform effect. Conditioning on
compliance --- restricting each format to the models that produce its wrapper at least 90\% of
the time, so incapacity stays out of the distributional signal --- the field-mean
$\Delta$-surprisal is $-0.22$ bits for JSON, $-0.23$ for XML, $-0.09$ for YAML, $-0.09$ for CSV,
and $+0.13$ for brackets. The two formats that compress the field are the two models are trained
to \emph{answer} in: JSON and XML, the registers of function calls, tool use, and
structured-output modes. The data formats models mostly only \emph{read} --- YAML and CSV --- do
not reliably compress it, and an arbitrary non-data wrapper (square brackets) \emph{loosens} it.

\paragraph{Significance.}
We test both exchangeable units with a sign-flip permutation (20{,}000 draws): field entropy
paired over the 31 categories, and within-column surprisal paired over the 44 models. JSON and
XML compress on both units (entropy $-0.20$, $p=.0006$ and $-0.20$, $p=.0004$; per model
$-0.22$, $p=.0002$ and $-0.19$, $p=.002$). YAML and CSV are not significant on either unit
($p=.75$ and $.46$ by category; $.80$ and $.35$ by model). Brackets significantly \emph{loosens}
on both ($+0.12$ entropy, $p=.014$; $+0.13$ per model, $p=.009$), and the loosening survives
every echo-guard variant --- $+0.16$ (no guard) through $+0.13$ (full guard) per model, $p<.01$
throughout --- so it is a real reversal, not a hygiene artifact. Why an arbitrary wrapper should
\emph{widen} the field we can only speculate: a bare \texttt{[answer]} slot reads less like a
data record than a fill-in-the-blank, a framing that may invite a more playful completion than
prose does; we report the reversal as a settled effect without a settled mechanism.
The JSON-versus-YAML difference, paired by
category, is itself significant ($p<10^{-4}$), and \textsc{deepseek-v3.2}'s individual collapse
clears $p<10^{-4}$ alone. The compression is thus specific to the trained answer-delivery
formats, which moves the weight of the mechanism toward tool-use post-training rather than a
generic property of serialized text. The gradient is not an artifact of the differing per-format
compliance subsets (compliant $n$: JSON 43, XML 41, YAML 37, CSV 39, brackets 43 of 44): on the
34 models compliant in \emph{all} five formats it reproduces and sharpens (JSON $-0.27$, XML
$-0.26$, brackets $+0.13$), while YAML and CSV stay weakly negative and non-significant.

\paragraph{But not post-training alone.}
The corpus-register account is not dead, because on the unconstrained ``Pick a word'' prompt
\emph{every} serialization concentrates the pool, including the two with no net battery effect:
\emph{serendipity} rises from 41\% in plain chat to 52\% (CSV), 59\% (XML), 64\% (JSON), and
65\% (YAML), while brackets holds at 41\%. YAML's single-category concentration is real; it
simply washes out across the full battery, which is why its net effect is null. A black-box
study cannot fully separate what the training corpus taught --- that text inside a data
structure is the canonical value --- from what tool-use tuning reflexively reinforced; our
reading is that both are present and the post-training component dominates the battery-wide
gradient.

\paragraph{Two companion phenomena.}
Compliance is not a nuisance to discard but a control that must be applied. \emph{Format
incompetence} masquerades as divergence: \textsc{granite} and \textsc{mythomax} emit a valid CSV
wrapper only about 1\% of the time, and their unwrapped replies, scored naively, read as high
surprisal --- \textsc{granite} shows a spurious $+1.49$ bits in CSV before conditioning. This is
why every distributional figure above is compliance-conditioned. Distinct from incompetence, a
fully-compliant model can still carry a positive $\Delta$ that is real signal rather than a parse
artifact: \textsc{llama-4-maverick} is 100\% compliant in JSON and CSV yet sits $+0.4$ to $+0.7$
bits above the field in every format --- not format failure but the \emph{stranding} described
below, the field moving while the model holds still. Register reaction is heterogeneous, not
uniformly compressive.

\paragraph{Per-model deltas: three tiers.}
Applying the census's tiers-not-ranks discipline to the deltas: (a) exactly six of 44 models
have an individually significant $\Delta$ at BH-FDR $q=.10$, and all six are compressions ---
\textsc{deepseek-v3.2} $-1.31$, \textsc{gpt-4o-mini} $-0.92$, \textsc{gpt-4-turbo} $-0.87$,
\textsc{gpt-5.6-sol} $-0.65$, \textsc{qwen3} $-0.45$, \textsc{gemini-3.1-pro} $-0.31$
(\textsc{hermes}, $-0.91$, falls just short --- its within-model dispersion is the panel's
largest (self-distinctness $0.71$, \S\ref{sec:temperature}), which widens its interval enough that
a near-identical delta misses the threshold). (b) The remaining ${\sim}38$ models --- including every model
with a positive $\Delta$ --- are individually indistinguishable from zero, and their mid-table
ordering carries no information; we do not interpret it. 

\paragraph{Divergence is often stranding, not motion.}
Because surprisal is scored within-column against the pool, a model that keeps its answers while
the field converges around it earns a rising $\Delta$ with no change in its own behavior ---
passive \emph{stranding}, not active divergence. A panel-free check separates the two: for each
model we compare its own chat answer distribution to its own JSON distribution (mean
Jensen--Shannon over the 31 categories, never touching the pool). The panel's largest positive
delta, \textsc{llama-4-maverick} ($+0.56$, 100\% JSON-compliant), has a \emph{below}-median
self-divergence (JSD $0.20$ vs the field's $0.27$; 7 of 31 modal answers change): it barely moves
its own answers, and its rising score is the collapsing field stranding a fixed, distinctive
point. Register-invariance is thus a model trait --- some models code-switch under serialization,
some hold their answers regardless --- and it is what most of the amber tail is. The trait is not
lineage-clean (\textsc{mythomax}, $+0.43$, \emph{does} move its own answers), so we read the
positive tail as a mix of stranding and genuine divergence the panel-relative $\Delta$ cannot
separate, and decline to rank it.

\subsection{Enforcement adds little the request did not}
\label{sec:enforced}
A natural objection is that the collapse is a decoding artifact. It is not: enforcing the schema
at the decoder compresses no further than asking for it. We ran one more column --- the JSON
clause plus a strict \texttt{response\_format} json\_schema --- on the 36 of 44 models whose
providers support it (the other eight return no valid output, a feature-acquisition signal in
itself). Field-mean surprisal --- plain, request, and enforcement all recomputed on the
36-model supported subset and scored within that subset's own pool, so the three are comparable ---
is 1.56 bits under the request and 1.53 under enforcement, against
1.79 in open chat: the request does the $-0.22$-bit work and enforcement adds $-0.03$ more. The
effect lives in the model's response to the register, not in the sampler. Per-model reactions are
heterogeneous and confounded --- \texttt{response\_format} is a native decoder constraint for
some providers and a gateway coercion for others --- so we read only the battery-wide mean and
release the data for a native-only replication.

\section{Limitations}
\label{sec:limitations}
Several limitations bound these claims. The data are a single snapshot served through one
channel (OpenRouter), and requested temperature 1.0 is not honored uniformly across providers
--- which is why we report self-distinctness alongside surprisal and rest the personality claims
on discrete four-of-four behavior. Surprisal is panel-relative: a model's bits depend on the
field it is scored against, so the absolute numbers are not portable across panels, although the
within-column and $\Delta$ comparisons are internal to this one. Each format is probed with a
\emph{single} clause wording, so we cannot separate the register from the particular phrasing
that invokes it; a clause-paraphrase column is the natural control. The gradient does, however,
already rule out the most obvious phrasing confounds: all four serialization clauses embed the
same \texttt{word} slot name and a fill-in template, yet only JSON and XML compress, so neither
key-priming nor the fill-in framing can be doing the work; and the one residual axis, clause
length, runs the wrong way for a length account --- CSV carries the wordiest clause and is null. Finally, we vary the register
through prompt-level \emph{requests} and one enforced \texttt{response\_format} counterpart
(\S\ref{sec:enforced}); tool-call framing and other structured-output pathways remain unmeasured,
and enforcement is itself served heterogeneously across providers, so a provider-controlled
native-enforcement replication would sharpen the per-model picture.

\section{Discussion}
\label{sec:discussion}
The result has one blunt practical consequence. Models are evaluated, compared, ranked, and felt
out in chat --- leaderboards, vibe checks, and human preference are all collected on the prose
surface --- but software consumes them through structured output, and that surface is measurably
more collapsed. Every diversity number the census reported is, for the deployment tier that
increasingly matters, an overstatement: the model an agent consults has narrower tastes than the
one a user chatted with, and the gap is a fixed cost of the register rather than a tail risk.
Synthetic-survey, LLM-judge, and tool-selecting pipelines all read the JSON persona. This bites
where the task admits many acceptable answers --- surveys, recommendations, brainstorming,
judgement, tool choice --- not where structured output carries a single correct value
(extraction, classification, routing), for which reduced diversity is no cost.

The register is also a cheap instrument in its own right, and format compliance falls out for free as a generational
capability track --- older models cannot speak some of these registers at all, and that
acquisition has a date a re-run will record. If the battery-wide compression we attribute mostly
to tool-use post-training is being trained in per release, a fixed public instrument run on
every model is what would detect it --- the same argument the census makes for measuring
conformity over time, now pointed at the register. The immediate
next columns are a provider-controlled native-enforcement replication and tool-call framing
(\S\ref{sec:enforced}) and clause paraphrases.

\section*{Data and code availability}
All code and data are in the \texttt{studies/structured/} directory of the modelun repository:
\texttt{run\_formats.py} (the battery runner), \texttt{analyze.py} (the metric, junk guard, and
$\Delta$-surprisal), \texttt{probe\_significance.py} (the permutation tests),
\texttt{analyze\_rtm\_splithalf.py} (split-half reliability and the regression-to-the-mean null),
\texttt{analyze\_register\_invariance.py} (the panel-free self-divergence check), and
\texttt{analyze\_brackets\_robustness.py} (the echo-guard sweep), against
\texttt{probes/format\_register.json} (the raw replies for all six columns). The plain-chat
baseline is the census transcripts; this study scores against the 44-model panel. An interactive
explorer --- scorecard, per-category and per-format tables, per-model drill-downs, and a
compliance tab --- is published alongside.

\section*{Note on AI usage}
This work was done in collaboration with Claude (Opus 4.8 and
Fable 5), which helped run the battery, build the analysis, and draft the text; the research
questions and interpretation are the author's. Both models are also subjects of the study.

\bibliographystyle{plainnat}
\bibliography{references}

\appendix
\section{Prompts}
\label{sec:prompts}
The battery is 31 single-turn prompts, frozen before data collection --- the census stimulus
unchanged. Thirty have the form \emph{``Name a[n] $X$.''} and the thirty-first removes the
category constraint (\emph{``Pick a word.''}). Each is followed by \emph{``Reply with one word
only.''} in plain chat, which in a format column is replaced by that format's clause
(\S\ref{sec:method}), with no other text changing. There is no system prompt, and each prompt is
a fresh single-turn conversation at requested temperature 1.0.

\begin{multicols}{2}
\small
\begin{itemize}\setlength{\itemsep}{1pt}
\item Name a color.
\item Name an animal.
\item Name a fruit.
\item Name a vegetable.
\item Name a city.
\item Name a country.
\item Name a flower.
\item Name a sport.
\item Name a musical instrument.
\item Name a bird.
\item Name a gemstone.
\item Name a tree.
\item Name a beverage.
\item Name an insect.
\item Name an occupation.
\item Name a language.
\item Name a dessert.
\item Name a tool.
\item Name a fish.
\item Name a metal.
\item Name a fabric.
\item Name an herb.
\item Name a dance.
\item Name a hobby.
\item Name a condiment.
\item Name a cheese.
\item Name a dinosaur.
\item Name a mythical creature.
\item Name a board game.
\item Name an emotion.
\item Pick a word.
\end{itemize}
\end{multicols}

\section{Normalization and junk guard}
\label{sec:norm}
Each reply is reduced to one answer token, following the census. For a format column the wrapper
is first stripped by the per-format regular expression (\S\ref{sec:hygiene}); the extracted
string --- or the raw reply if no wrapper matches --- is then normalized: lowercased, stripped of
surrounding punctuation and emoji, and reduced to its final alphabetic word (so \emph{``A common
color is blue.''} $\to$ \emph{blue}). A mechanical junk guard treats the following as \emph{failed
cells} rather than answers: chat-template artifacts (\eg \texttt{[/INST]} or markup tags), replies
longer than 15 words (a truncated chain of thought would otherwise contribute a spurious novel
final word), bare acknowledgements (\emph{``Okay.''}), and single-letter tokens. Within a
category, bare plurals are merged with their singulars when both occur. This is the identical rule
set used by the census, imported rather than re-implemented so the two studies cannot drift. The
format columns add one guard the plain census does not need: a reply whose normalized token is
the category noun or the fill-in placeholder (\texttt{[city]}, \texttt{answer}, \texttt{word}) is
an \emph{echo} of the wrapper's own slot rather than an answer, and is treated as a failed cell.
The guard runs on every column alike --- it removes nothing from plain chat, which has no slot to
echo --- and its only material effect is to remove the echo mass a fill-in wrapper elicits (most
of it in CSV and brackets), which trims the apparent brackets loosening from $+0.16$ to $+0.13$
bits and is why the raw and compliance-conditioned brackets deltas agree after it is applied.

\end{document}